\documentclass[preprint,review,12pt]{elsarticle}

\usepackage{amssymb, amsmath, amsthm, bm}
\usepackage[noend]{algpseudocode}
\usepackage{algorithmicx,algorithm}
\usepackage{graphicx}
\usepackage[caption=false]{subfig}
\usepackage[colorlinks,linkcolor=blue]{hyperref}
\usepackage{epstopdf}
\usepackage{diagbox}

\journal{}

\begin{document}
	
\begin{frontmatter}
		
\title{PC-GAIN: Pseudo-label Conditional Generative Adversarial Imputation Networks for Incomplete Data}

\author[label1] {Yufeng Wang \fnref{cor1}}
\author[label2] {Dan Li \fnref{cor2}}
\author[label3] {Xiang Li \fnref{cor3}}
\author[label1] {Min Yang \corref{cor4}}
\fntext[cor1] {Email: ytuyufengwang@163.com }
\fntext[cor2] {Email: lidan2017@ia.ac.cn}
\fntext[cor3] {Email:lixiang2020ecnu@163.com}
\cortext[cor4] {Corresponding author: yang@ytu.edu.cn}

\address[label1]{School of Mathematics and Information Sciences, Yantai University, Yantai, China}
\address[label2]{Research Center for Brain-inspired Intelligence and National Laboratory of Pattern Recognition,
Institute of Automation Chinese Academy of Sciences, Beijing, China}	
\address[label3]{Software Engineering Institute, East China Normal University, Shanghai, China}	
		
\begin{abstract}
Datasets with missing values are  very common in real world  applications.
GAIN, a recently proposed  deep generative model for missing data imputation,
has been proved to outperform many state-of-the-art methods.
But GAIN only uses a reconstruction loss in the
generator to minimize the imputation error of the non-missing part,
ignoring the potential category information which can reflect the relationship between samples.
In this paper, we propose a novel unsupervised missing data imputation method named PC-GAIN,
which utilizes potential category information to further enhance the imputation power.
Specifically, we first propose a pre-training procedure
to learn potential category information contained in a subset of low-missing-rate data.
Then an auxiliary classifier is determined using the synthetic pseudo-labels.
Further, this classifier is incorporated into the generative adversarial framework to help the generator to yield higher quality imputation results.
The proposed method can improve the imputation quality of GAIN significantly.
Experimental results on various benchmark datasets show that our method is also superior to other baseline approaches.
Our code is available at \url{https://github.com/WYu-Feng/pc-gain}.
\end{abstract}

\begin{keyword}
 conditional; generative adversarial network; imputation; missing data;  pseudo-label
\end{keyword}
		
\end{frontmatter}

\section{Introduction}
\begin{figure*}[!t]
\centering
\subfloat[Complete samples]{

%\begin{minipage}[t]{0.5\textwidth}
\centering
\includegraphics[width=1.75in,height=0.93in]{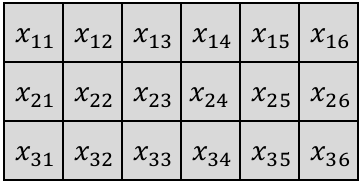}
%\end{minipage}
}
\hspace{15mm}
\subfloat[Incomplete samples]{
\label{}
%\begin{minipage}[t]{0.5\textwidth}
\centering
\includegraphics[width=1.75in,height=0.93in]{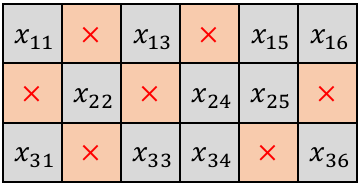}
%\end{minipage}
}
\caption{Examples of complete and incomplete data with dimension 6.}
\label{data}
\end{figure*}

Data analysis is a core component of scientific research across many domains  \cite{Abiri2019,Gondara2017,Kiranyaz2020,Mottini2018,Sevgen2017}.
Missing data can degrade model quality and even lead to incorrect insights \cite{Garica-Laencina2010,Little2019}.
If the number of incomplete samples (examples of complete and incomplete data are shown in Figure \ref{data}) is small, then we can drop them.
However, dropping too many samples may diminish the statistical power of subsequent analysis because of the lack of remaining data.
So an effective solution is to perform data imputation,
that is replacing missing values with estimated values.

\smallskip
There is extensive literature on missing data imputation.
These literature can be mainly classified into two categories:
discriminative models and generative ones.
Examples of discriminative models with state-of-art performance
include MICE \cite{Buuren2011}, MissForest \cite{Buhlmann2012}, and Matrix Completion \cite{Mazumder2009}.
Compared to discriminative models,
generative models are usually better in capturing complex nonlinear correlations in the missing data.
EM algorithm based on Gausssian mixture assumption \cite{Garica-Laencina2010}  is a classical generative model.
In recent years, advances in deep generative models have made it possible to significantly increase the quality of imputation results, see e.g.
\cite{Abiri2019,Gondara2017,Spinelli2020,Tran2017,Xu2018}.
In particular,
Yoon et al. \cite{Yoon2018} presented a generative adversarial imputation network (GAIN) for missing data imputation,
where the generator outputs a completed vector conditioned on what is actually observed,
and the discriminator attempts to determine which entries in the completed data were observed and which were imputed.
GAIN has been shown to outperform many state-of-the-art imputation models.
\smallskip
However, note that  in the framework of GAIN \cite{Yoon2018},
only a reconstruction loss is used in the generator to minimize the imputation error of the non-missing part.
It is well known that many real life datasets contain potential category information that has a close relationship with the latent feature distribution.
Incorporating such  information into the framework of GANs can improve the performance of the models  further \cite{Liu2020,Lucic2019,Sage2018}.

\smallskip
In this paper,
we aim to exploit the implicit category information contained in the incomplete data,
and develop a novel pseudo-label conditional GAN (PC-GAIN) to enhance the imputation quality of GAIN \cite{Yoon2018}.
The starting point of our method is simple.
Specifically, we first select a subset of low-missing-rate samples to perform a pre-training using  the original GAIN
and then synthesize their pseudo-labels by applying a clustering algorithm.
Using only a subset of low-missing-rate data is critical to ensure the quality of pseudo-labels
and has a crucial impact on the model's performance (Section 4.3).
Then, an auxiliary classifier is determined  based on these imputed samples and the corresponding pseudo-labels.
Further, the classifier is incorporated into the generative adversarial framework to
help the generator to yield indistinguishable imputation results,
while retaining  better category information.
We evaluate PC-GAIN  on the UCI and MNIST datasets with various missing rates.
Experimental results show that our PC-GAIN is superior to the baseline algorithms,
including GAIN,  especially when the missing rate is high.

The main contributions of this work are summarized as follows:
\begin{itemize}
\item [(1)]
 We propose a novel conditional GAN that exploits the implicit category information contained in the incomplete data
 to further enhance the imputation quality of GAIN \cite{Yoon2018}.
\item [(2)]
 We design an efficient pre-training procedure that only selects a part of low-missing-rate samples to perform imputation
 and  thus improve the quality of pseudo-labels.
 \item [(3)]
  An auxiliary classifier, along with the discriminator, is designed to help the generator
  to produce  indistinguishable imputation results, while preserving better category information.
\item [(4)]
 We show that our method outperforms state-of-the-art  methods in both the imputation and the prediction accuracy,
 especially when the missing rate is high.
 Moreover, regardless of the real number of categories,
 selecting a smaller cluster number can ensure the best performance of the model in practice,
 a property that makes the approach more flexible.
\end{itemize}

There are three types of missing mechanism \cite{Little2019}:
(1) missing completely at random (MCAR);
(2) missing at random (MAR), where the propensity for a data item to be missing is related to the observed data;
(3) missing not at random (MNAR), where the data items are missing due to some underlying mechanisms.
As in \cite{Mottini2018,Spinelli2020,Yoon2018}, we  consider the data that are MCAR in this paper.
This indicates that the missingness is caused by either unexpected external factors or control of the measurement system.

\section{Related Work}

\subsection{ Generative Adversarial Networks}
A large number of neural networks have been proposed for different problems in practical applications \cite{Gondara2017,Kiranyaz2020,Sevgen2017}.
Among them, generative adversarial networks (GANs) \cite{Goodfellow2014}, a framework to construct a generative model to approximate the target distribution,
have achieved state-of-the-art performance in various learning tasks \cite{Mottini2018,Mukherjee2019,ORTAC2020,Xu2018}.
The most significant feature of GANs is the discriminator which distinguishes the difference between the generated distribution and the target distribution.
The algorithm of GANs  iteratively trains the discriminator and generator,
where the discriminator acts as an increasingly rigorous criticism of the current generator.
However, the original GANs do not use additional information such as labels  to monitor the training,
and thus may lose important patterns in the generated results.

\smallskip
Conditional GANs (cGANs) \cite{Mirza2014} can be regarded as an augmentation of GANs
that use conditional information such as labels to improve the discriminator and generator.
Conditional GANs have been popularly used in conditional image synthesis \cite{Wang2018},
the generation of the images from text \cite{Reed2016}, and image to image translation \cite{Isola2017}.
Unlike  original GANs,
the discriminator of cGANs discriminates between the generator distribution and the target distribution on the set of generated samples and its conditional variable.
However, in practice, it is often expensive to obtain actual labels for large-scale datasets.
Therefore, for unlabeled or partially labeled data,
one solution is to modify the discriminator to predict the class distribution \cite{Odena2017,Springen2015}
or a subset of latent variables from which the samples are generated \cite{Chen2016}.
Another solution is to train class-conditional
GANs on unlabelled data by clustering on features obtained by unsupervised learning methods \cite{Liu2020,Lucic2019,Sage2018}.

\subsection{Deep Generative Imputation Methods}
Traditional  methods for missing data imputation usually assumes a linear relationship between the observed part and the missing part of the data,
However, such linear relationship usually does not exist in reality.
Because of the strong nonlinear fitting ability of neural networks,
there has been a surge of interest in developing deep generative models for missing data imputation.
For example, Gondara and Wang \cite{Gondara2017} considered a multiple imputation model based on overcomplete deep denoising autoencoders.
Tran et al. \cite{Tran2017} proposed a cascade residual automatic encoder(CRA) which compensates for missing data
by utilizing the correlation between different modalities.
Mattei \cite{Mattei2019} presented an importance-weighted autoencoder that maximises a potentially tight lower bound
of the log-likelihood of the observed data.
Nazabal et al. \cite{Nazabal2020} proposed a general framework of variational autoencoders
that effectively incorporates incomplete data and heterogenous observations.
Spinelli et al. \cite{Spinelli2020} formulated the missing data imputation task in terms of a graph denoising autoencoder,
where each edge of the graph encodes the similarity between two patterns.

\smallskip
Imputation methods using GAN frameworks have also been proposed.
Pathak et al. \cite{Pathak2016} presented context encoders using a pixel-wise reconstruction loss plus an adversarial loss
to generate the contents of an arbitrary image region conditioned on its surroundings.
Xu et al. \cite{Xu2018} presented a tabular GAN where the generator outputs variable values in an ordered
sequence using a recurrent neural network architecture.
Mottini  et al \cite{Mottini2018} handled PNRs data with missing values by use of a Cramer GAN \cite{Bellemare2018},
where feedforward layers with the Cross-Net architecture and an input embedding layer for the categorical features are added.
Yoon et al. \cite{Yoon2018} proposed a generative adversarial imputation network (GAIN) for missing data imputation,
where the generator outputs a completed vector conditioned on what is actually observed,
and the discriminator attempts to determine which entries in the completed data were observed and which were imputed.
Li et al. \cite{Li2019} introduced an auxiliary GAN for learning a mask distribution to model the missingness.
The complete data generator is trained so that the resulting masked data are indistinguishable
from real incomplete data that are masked similarly.
However, training two generative adversarial networks at the same time can be quite computationally complex.
Moreover, the generative  imputation models  mentioned above have not considered the use of
implicit category information to improve the imputation performance.

\section{Pseudo-label conditional GAIN}

\smallskip
Let $ \mathbf{\chi}=\{\mathbf{x}^1,\mathbf{x}^2,\cdots, \mathbf{x}^N\}  \in \mathbb{R}^d $ denote an incomplete dataset.
For each $ \mathbf{x} \in \mathbf{\chi}  $, there is a corresponding binary mask vector  $ \mathbf{m}= \{0,1\}^d $,
where $ m_i = 1 $ if the feature $ x_i $ is observed,  and $ m_i = 0 $ if $ x_i $ is missing.

\subsection{The Review of GAIN}

\smallskip
In \cite{Yoon2018},
the authors proposed a generative adversarial neural network called GAIN to impute the missing values.
In GAIN, the generator $ G $ takes the incomplete sample $ \mathbf{x} $, the mask vector  $ \mathbf{m} $
and a source of noise as input and output the complete sample,
and then the discriminator $ D $  tries to distinguish which entries are observed and which are imputed.
Further, to alleviate the diversity of the solution,
a hint mechanism was  introduced to provide additional missing  information for the discriminator.

Specifically, the output of the generator $ G $ in GAIN could be denoted by
\begin{align}
  \mathbf{x}_{G} = G(\mathbf{x}, \mathbf{m}, (1-\mathbf{m}) \odot \mathbf{z}),
\end{align}
where $ \mathbf{z} $ is a $d $-dimensional noise and  $\odot$ denotes the Hadamard product.
The reconstructed sample is defined as
\begin{align}
x_{R} = \mathbf{m}  \odot \mathbf{x}  + (1-\mathbf{m}) \odot  \mathbf{x}_G.
\end{align}
The output of the discriminator  $ D $ is an binary vector denoted by
\begin{align}
    \mathbf{m}_{D} = D(\mathbf{x}_R,\mathbf{h}),
\end{align}
where $ \mathbf{h} $ is a hint vector and  $ \mathbf{m}_{D} $ is the prediction of the mask vector $ \mathbf{m} $.

\smallskip
The objectives of GAIN are formulated as follows
\begin{align}
\label{GAINObj}
\begin{split}
     \min_{D}&\frac{1}{N}\sum_{k=1}^{N}\mathcal{L}_{D}(\mathbf{m}^k,\mathbf{m}_{D}^k),
     \\[5pt]
     \min_{G}&\frac{1}{N}\sum_{k=1}^{N}\left( \mathcal{L}_{G}(\mathbf{m}^k,\mathbf{m}_{D}^k)
     + \alpha \mathcal{L}_{R}(\mathbf{x}^k,\mathbf{x}_R^k) \right),
\end{split}
\end{align}
where $ \alpha $ is a weight parameter,
$ \mathcal{L}_{D} $ is a cross entropy loss term such that
\begin{align}
\mathcal{L}_{D}(\mathbf{m},\mathbf{m}_{D}) = - \mathbf{m} \log \mathbf{m}_{D} - (1-\mathbf{m})\log(1-\mathbf{m}_{D}),
\end{align}
$ \mathcal{L}_G $ is defined as
\begin{align}
\mathcal{L}_{G}(\mathbf{m},\mathbf{m}_{D}) = -(1-\mathbf{m}) \log \mathbf{m}_{D},
\end{align}
and $  \mathcal{L}_{R} $ is a reconstruction loss  satisfying
\begin{align}
     \mathcal{L}_{R}(\mathbf{x},\mathbf{x}_R) =\sum_{i=1}^d m_i   L_{R}(x_i,x_{R,i}),
\end{align}
with
\begin{align}
  L_{R}(x_i,x_{R,i})=
\left \{
      \begin{array}{ll}
             (x_i-x_{R,i})^2, & \text{for numerical variable}  \\[5pt]
            -x_i \log x_{R,i}, & \text{for categorial variable}
\end{array} \right..
\end{align}

\subsection{ Formulation of PC-GAIN}

It is well known that the conditional information such as labels can enhance the performance of the generator \cite{Liu2020,Mirza2014}.
However, applying the existing conditional techniques  to common imputation problems mainly faces two difficulties.
First, most imputation problems are completely unsupervised  and there are no explicit labels could be used directly.
Secondly, since the data are incomplete,  it is particularly difficult to synthesize appropriate pseudo-labels for the samples.

\begin{figure}[hbt]
\centering
\includegraphics[width=5.5in,height=3.9in]{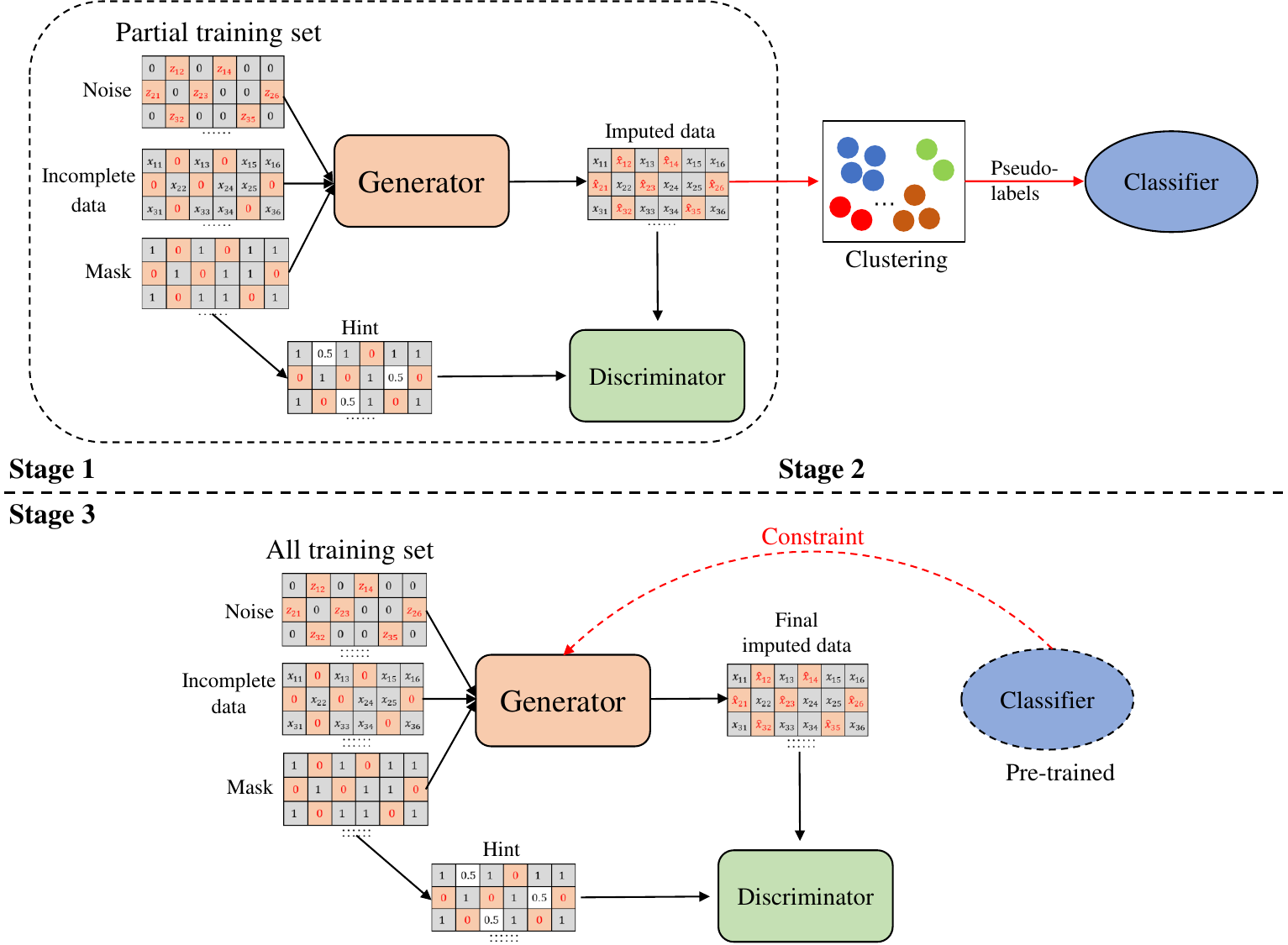}
\caption{An overview of the proposed PC-GAIN.}
\label{framework}
\end{figure}

In this section,
we would like to propose a novel algorithm based on GAIN to solve the above difficulties.
Figure \ref{framework} shows the whole framework of the proposed method.
We first select low-missing-rate samples to pre-train the generator $ G $ and the discriminator $ D $ to obtain the imputed dataset.
Then, a clustering algorithm is applied on the imputed dataset to synthesize the pseudo-labels.
And we train a classifier with imputed dataset and pseudo-labels.
Finally, we use all training data to train the generator $ G $  and the discriminator $ D $,
and simultaneously use the pre-trained classifier to constrain the generator.
The details of the proposed method are described in the following.

\smallskip
First, note that even under a fixed missing rate,
the missingness of each sample in a dataset is different.
In our opinion, the potential category information  contained in those low-missing-rate samples is more reliable.
Therefore, we would like to select a subset of low-missing-rate  samples to conduct a pre-training procedure.
The aim of the pre-training is to impute the missing components of these low-missing-rate data and then deduce their pseudo-labels.

\smallskip
Specifically, for any $ \mathbf{x} $, we calculate its missing rate $ r(\mathbf{x}) $ as follows
\begin{align}
 r(\mathbf{x}) = \frac{1}{d} \sum_{i=1}^{d} m_{i}.
\end{align}
where $ \mathbf{m} $ is the mask vector of the data.
Then we sort all samples in an ascending order according to their missing rates,
and choose the first $ \lambda N $ ( $ 0<\lambda<1 $)  samples to formulate a pre-training dataset  $\chi^{L}$.
Following the optimization objectives \eqref{GAINObj},
we pre-train the generator $ G $ and the discriminator $ D $ using the dataset  $\chi^{L}$.
After then, we can obtain an imputed dataset denoted as $\chi_R^{L}$ of the dataset $\chi^{L}$.

\smallskip
To synthesize the pseudo-labels $ \{\mathbf{p}_R^L\} $ of these low-missing-rate data,
we can apply a clustering algorithm, e.g. \cite{Arthur2007,Lucic2019,Mukherjee2019}, on the imputed dataset $\chi_R^{L}$.
It is worth to point out that the number of the clusters is not need to be consistent with the number of the real categories.
In our experiments,
we found that for many UCI datasets,
a small value of $ K $  between 4 and 6 is enough to ensure the best performance of the model.

\smallskip
Next, we use $ \chi_R^{L}=\{\mathbf{x}_R^L\} $
and the corresponding pseudo-labels $ \{\mathbf{p}_R^L\} $ to train an auxiliary classifier $ C $.
With the aid of this classifier,
we update the generator $ G $ and discriminator $ D $ again.
That is, we require the generator not only outputs indistinguishable imputed data,
but also learns distinct categorical characteristics.
More specifically,
the objectives of the discriminator and the generator now become
\begin{align}
\label{PCGAINObj}
\begin{split}
     \min_{D}&\frac{1}{N}\sum_{k=1}^{N}\mathcal{L}_{D}(\mathbf{m}^k,\mathbf{m}_{D}^k),
     \\[5pt]
     \min_{G}&\frac{1}{N}\sum_{k=1}^{N}\left( \mathcal{L}_{G}(\mathbf{m}^k,\mathbf{m}_{D}^k)
     + \alpha \mathcal{L}_{R}(\mathbf{x}^k,\mathbf{x}_R^k)
     +\beta \mathcal{L}_{C}(\mathbf{x}_R^k) \right),
\end{split}
\end{align}
where $ \alpha $ and $ \beta $ are hyperparameters,  and $ L_C $ is  a standard information entropy loss such that
\begin{align}
\mathcal{L}_{C}(\mathbf{x}_R) = -C(\mathbf{x}_R) \log C(\mathbf{x}_R),
\end{align}
with $ C(\mathbf{x}_R) $ denoting the output of the auxiliary classifier.

\smallskip
Note that the main difference between \eqref{GAINObj} and \eqref{PCGAINObj} is that
in the objectives of PC-GAIN,
an additional entropy loss $ L_C $ is introduced to promote the model to learn more distinct categorical features.
This loss is determined by the auxiliary classifier $ C $, which is only pre-trained on a subset $ \{\mathbf{x}_R^L,\mathbf{p}_R^L\} $,
and fixed during the training of the generative adversarial network.
Hence, this additional classier will not increase the training complexity of the GANs.
The pseudo-code for our PC-GAIN is shown in Algorithm 1.

\makeatletter
\def\BState{\State\hskip-\ALG@thistlm}
\makeatother
\begin{algorithm}
\label{Algorithm PCGAIN}

\caption{Pseudo-Code of PC-GAIN}
{\bf Input:} Incomplete dataset $ \chi $,  number of clusters $ K $ and proportion parameter $ \lambda  $.

{\bf Pre-training}
\begin{algorithmic}[1]
	\State Sort all data in an ascending  manner according to their missing rates.
	\State Select the top $ \lambda $ ($0< \lambda <1$) of the data to formulate a pre-training dataset  $\chi^{L}$.
    \State Use $ \chi^L $ and the objectives \eqref{GAINObj}  to train the discriminator $ D $ and the generator $ G $.
\end{algorithmic}

{\bf Determine the pseudo-labels and the classifier $ C $}
\begin{algorithmic}[1]
   \State Cluster the imputed results of the first stage and synthesize the pseudo-labels for these data.
   \State Train a classifier  $ C $  using the imputed data and their pseudo-labels.
\end{algorithmic}

{\bf Updata the generator $ G $ and the discriminator $ D $}
\begin{algorithmic}[1]
   \State Use the whole dataset $ \chi $ and the new objectives \eqref{PCGAINObj}  to train the discriminator $ D $ and the generator $ G $ again. 	
\end{algorithmic}

\end{algorithm}

\section{ Experiments}
\subsection{Experimental Setup}
\subsubsection{Datasets}
In this section, we evaluate the performance of PC-GAIN on six datasets from the UCI repository \cite{Dua2017}
and the MNIST dataset  \cite{LeCun2010}, respectively.
The details of the UCI datasets are listed in Table \ref{UCI}.

\begin{table}[htb]
\center
\caption{The basic properties of the UCI datasets}
\label{UCI}
\scalebox{0.85}{
     \begin{tabular}{|c|c|c|c|c|}
     \hline
	 Dataset                  	&	Samples	&	Numerical variables	&	Categorial variables	&	Number of classes	\\
	\hline									
	BalanceScale			&	625		&	0				&	4				&	3				\\	%	dataset from http://archive.ics.uci.edu/ml/datasets/Balance+Scale
	BreastCancer		&	{569}	&	{30}				&	{0}				&	{2}				\\
	CarEvaluation		    	&	1728		&	0				&	6				&	4				\\	%	dataset from http://archive.ics.uci.edu/ml/datasets/Car+Evaluation
	{Credit}				&	{30000}		&	{14}				&	{9}				&	{2}				\\
	Letter				&	20000		&	16	    			&	0				&	26				\\	%	dataset from https://archive.ics.uci.edu/ml/datasets/Letter+Recognition
	News					&	39797		&	35	   			&	23	    			&	2				\\	%	dataset from https://archive.ics.uci.edu/ml/datasets/Online+News+Popularity
	Spam					&	4601		&	57  				&	0				&	2				\\	%	dataset from https://archive.ics.uci.edu/ml/datasets/Spambase
	WineQuality(white)		&	4898		&	11  				&	0				&	7				\\	%	dataset from https://archive.ics.uci.edu/ml/datasets/Wine+Quality		
	\hline
\end{tabular}}
\end{table}

Since no dataset contains missing values originally,
for a given missing rate,
we remove the features of all data completely at random to formulate an incomplete dataset.
Moreover, each variable is scaled to the interval  $ [0,1] $.

\subsubsection{Compared Methods and Experimental Settings}
We shall compare PC-GAIN  with a range of baseline methods,
including Autoencoder \cite{Gondara2017},
EM \cite{Garica-Laencina2010},
MissForest \cite{Buhlmann2012},
MICE \cite{Buuren2011} and GAIN \cite{Yoon2018}.
Among them, MissForest and MICE belong to classical benchmark discriminative models,
while Autoencoder, EM and GAIN are generative methods.
To make a fair comparison of PC-GAIN and GAIN,
we adopt the same generative adversarial network architecture as \cite{Yoon2018}.
We take 5-cross validation in our experiments.
Each experiment is repeated ten times and the average performance is reported.

\smallskip
Unless stated otherwise,
the performance of each model is evaluated under a 50\% missing rate,
the weight parameters in \eqref{PCGAINObj} are chosen as $ \alpha=200 $ and $ \beta=20 $,
and the number of clusters $ K =5 $.
We use $ \lambda= 0.2 $ for three bigger datasets: Credit, Letter and News,
and $ \lambda= 0.4 $ for others.

\smallskip
In our PC-GAIN,  the auxiliary classifier is a three-layer  fully connected  neural network (NN) using ReLU activation.
The number of neurons in each hidden layer is the same as the input dimension of the data.
We apply the  KMeans++ clustering  \cite{Arthur2007} on the original imputation results of the UCI data,
while for the MNIST dataset, we apply KMeans++ to cluster the latent feature space of the generator.

\subsection{ Imputation Accuracy in UCI Datasets}
RMSE is a commonly used metric for evaluating the performance of
missing data imputation, which computes the root mean square error of the imputed missing values against the
ground truth.

\begin{table}[!htb]
\center
\caption{Comparisons with state-of-the-art methods on the UCI datasets. RMSE under a 50\%  missing rate. }
\label{RMSE50}
\scalebox{0.7}
{\begin{tabular}{|c|c|c|c|c|c|c|c|c|c|}
     \hline
 \diagbox{Method}{Dataset}                   	&	BalanceScale	&	{BreastCancer}	&	CarEvaluation  	&	{Credit}			& 	Letter  		& 	News 			&	Spam       	&	WineQuality	\\		
\hline																	
Autoencoder \cite{Gondara2017}  			&	0.4788		&	{0.1148}		&	0.4814		&	{0.1932}			&	0.2216		&	0.2726		&	0.0957		&	0.1814		\\		
EM	\cite{Garica-Laencina2010}		        	&	0.4992		&	{0.1176}		&	0.5492		&	{0.1923	}		&	0.2183		&	0.3064		&	0.0737		&	0.1520		\\		
MissForest	\cite{Buhlmann2012}	    			&	0.4322		&	{0.1073}		&	0.4784		&	{0.1817	}		&	0.1905		&	0.2633		&	0.0673		&	0.1597	    \\		
MICE	\cite{Buuren2011}		       		&	0.5161		&	{0.1158}		&	0.5521		&	{0.1843}			&	0.1714		&	0.2763		&	0.1062		&	\textbf{0.1304}	\\		
GAIN	 \cite{Yoon2018}		        		&	0.4148		&	{0.0944}		&	0.4343		&	{0.1832}			&	0.1587		&	0.2532		&	0.0583		&	0.1397		\\		
PC-GAIN        							&	\textbf{0.3754}	&	{\textbf{0.0939}}	&	\textbf{0.4227}	&	{\textbf{0.1616}}		&	\textbf{0.1450}	&	\textbf{0.2269}	&	\textbf{0.0559}	&	0.1323	\\		
\hline
\end{tabular}}

\end{table}

We first report RMSE of PC-GAIN and the baseline models in Table \ref{RMSE50}.
It can be seen that, except that it is slightly worse than the MICE method on the WineQuality dataset,
the performance of PC-GAIN outperforms the benchmark methods on most datasets.

\begin{figure}[hbt]
\centering
\subfloat[News]{\label{fig:mdleft}{\includegraphics[width=0.33\textwidth]{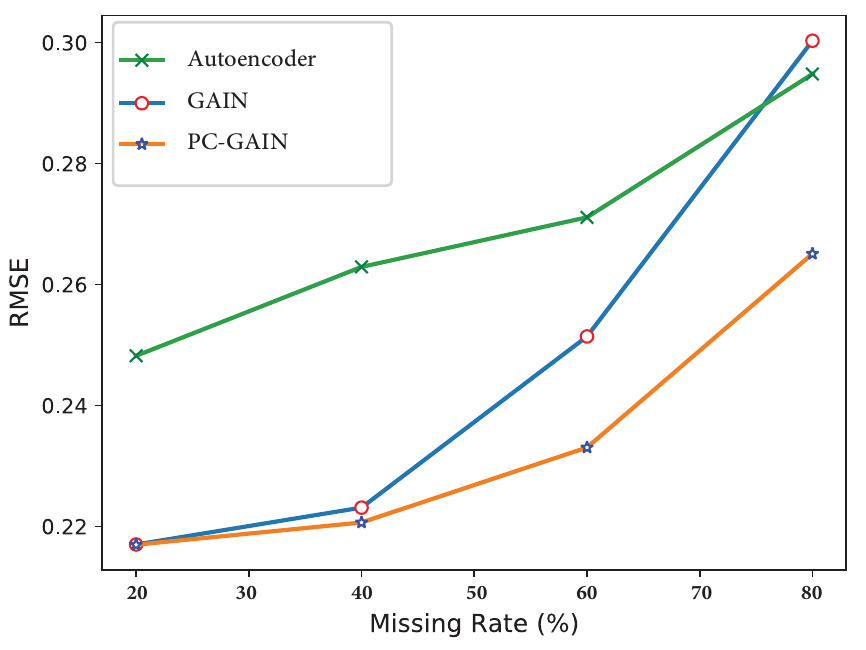}}}\hfill
\subfloat[Spam]{\label{fig:mdright}{\includegraphics[width=0.33\textwidth]{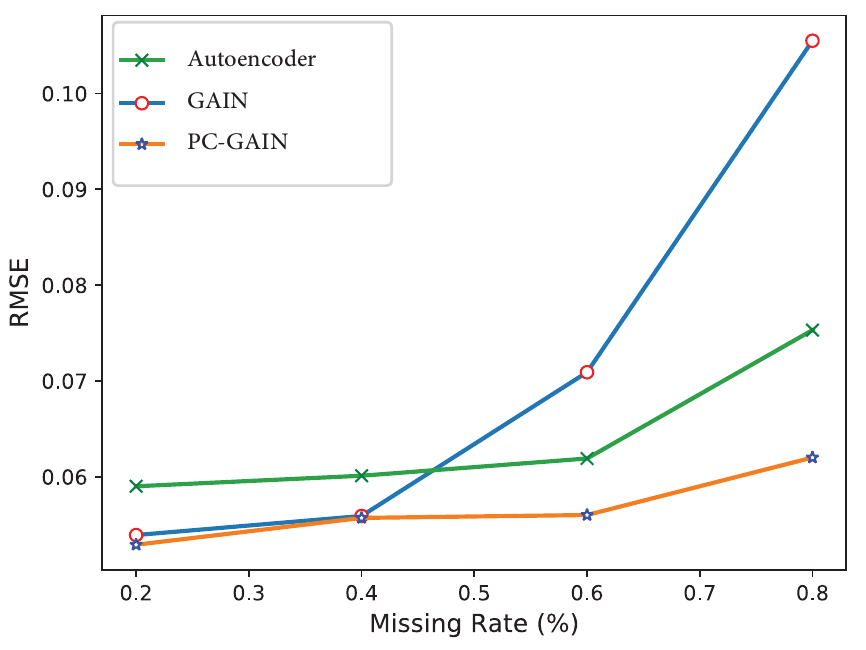}}}\hfill
\subfloat[Winequality]{\label{fig:mdright}{\includegraphics[width=0.33\textwidth]{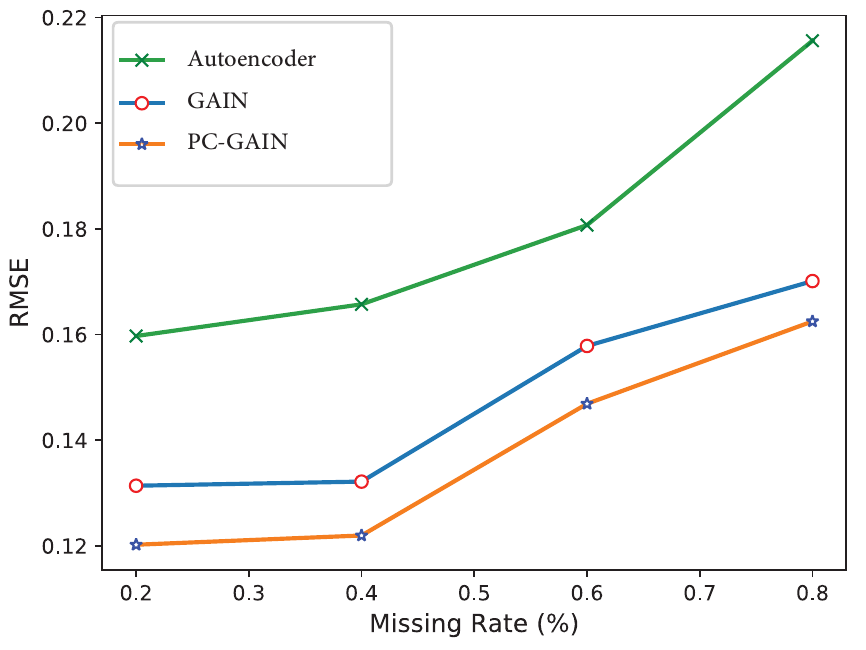}}}
\caption{ RMSE of PC-GAIN versus AutoEncoder and GAIN  with various missing rates.}
\label{RMSEVari}
\end{figure}

\smallskip
Next, we use News, Spam and WineQualy as examples
to illustrate the performance of PC-GAIN under various missingness.
As shown in Figure \ref{RMSEVari},
PC-GAIN consistently outperforms the compared models under  various missingness.
Especially, the advantage becomes more obvious when the missing rate becomes higher.

\subsection{Prediction Performance under Various Missing Rates}
It is a common understanding that a good imputation model should not only restore the data accurately,
but also maintain the category information of the data.
In this section,
we compare PC-GAIN against  other two deep generative models: Autoencoder \cite{Gondara2017} and GAIN \cite{Yoon2018},
with respect to the accuracy of post-imputation prediction.
We use a same classifier (a two-layer fully connected network with Softmax activation) for all three modes.

\begin{figure}[hbt]
\centering
\subfloat[News]{\label{fig:mdleft}{\includegraphics[width=0.333\textwidth]{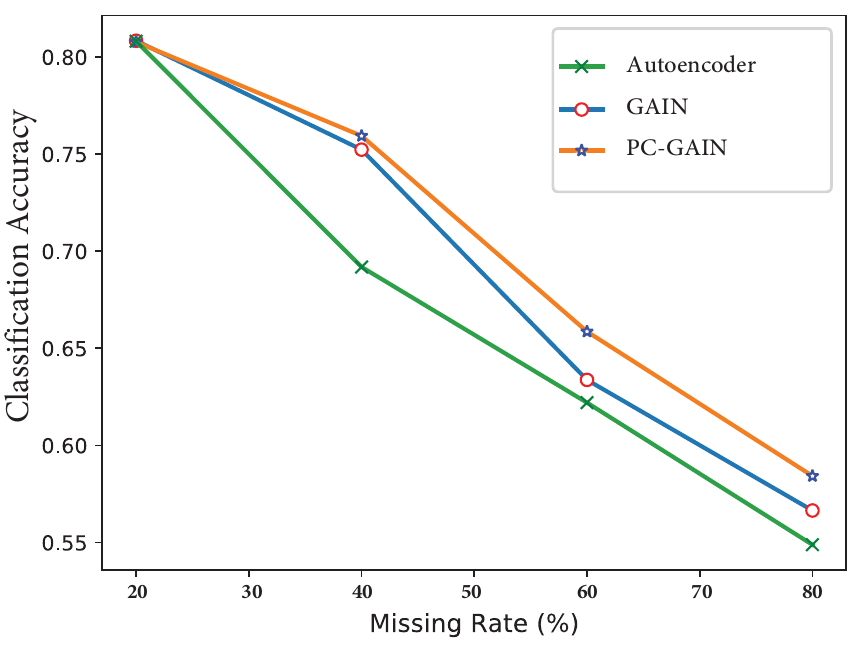}}}\hfill
\subfloat[Spam]{\label{fig:mdright}{\includegraphics[width=0.333\textwidth]{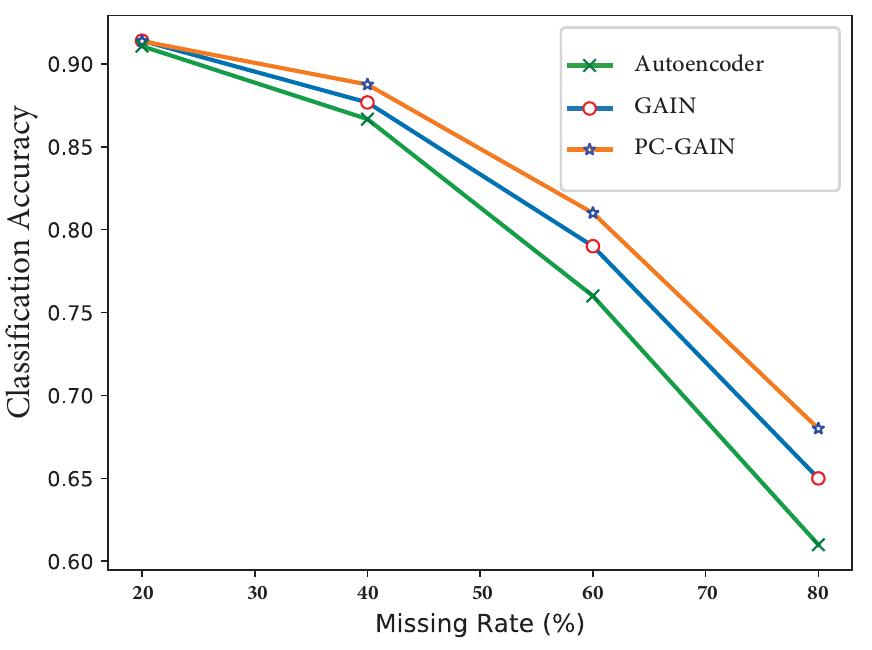}}}\hfill
\subfloat[Winequality]{\label{fig:mdleft}{\includegraphics[width=0.333\textwidth]{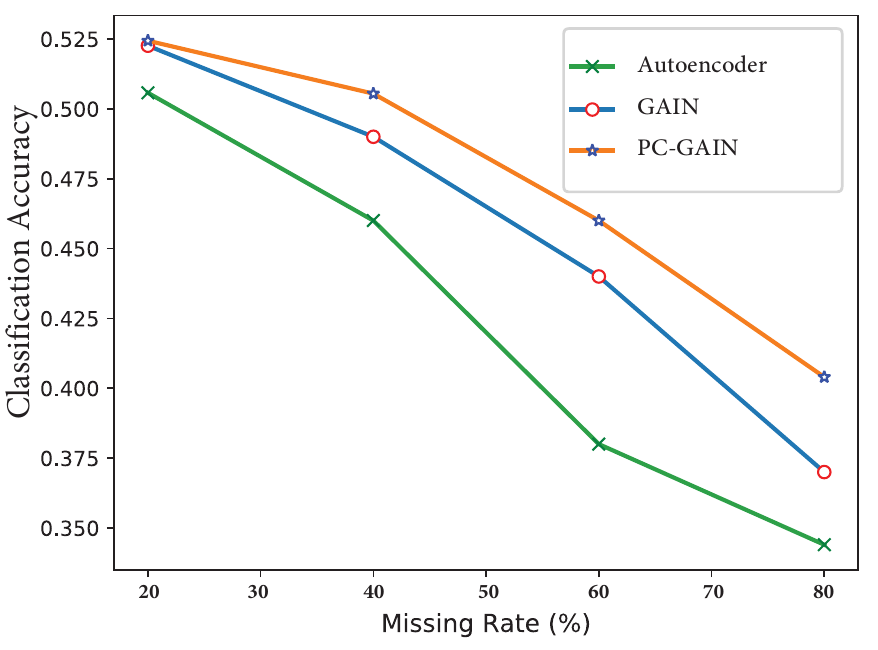}}}
\caption{Post-imputation prediction accuracy of PC-GAIN versus AutoEncoder and GAIN with various missing rates.}
\label{ClassificationAcc}
\end{figure}

\smallskip
It is observed from Figure \ref{ClassificationAcc} that PC-GAIN achieves the best classification accuracy in all cases.
Moreover,  the improvements in prediction accuracy  become more significant as the missing rate increases.
This phenomenon, with the imputation accuracy shown in Figure \ref{RMSEVari},
 implies that PC-GAIN is a reliable imputation method especially in the case of high missing rate.

\subsection{Sensitivity Analysis}
This section evaluates the performance of  PC-GAIN under various configurations.

\begin{figure}[hbt]
\centering
\subfloat[Spam]{\label{fig:mdleft}{\includegraphics[width=0.4\textwidth]{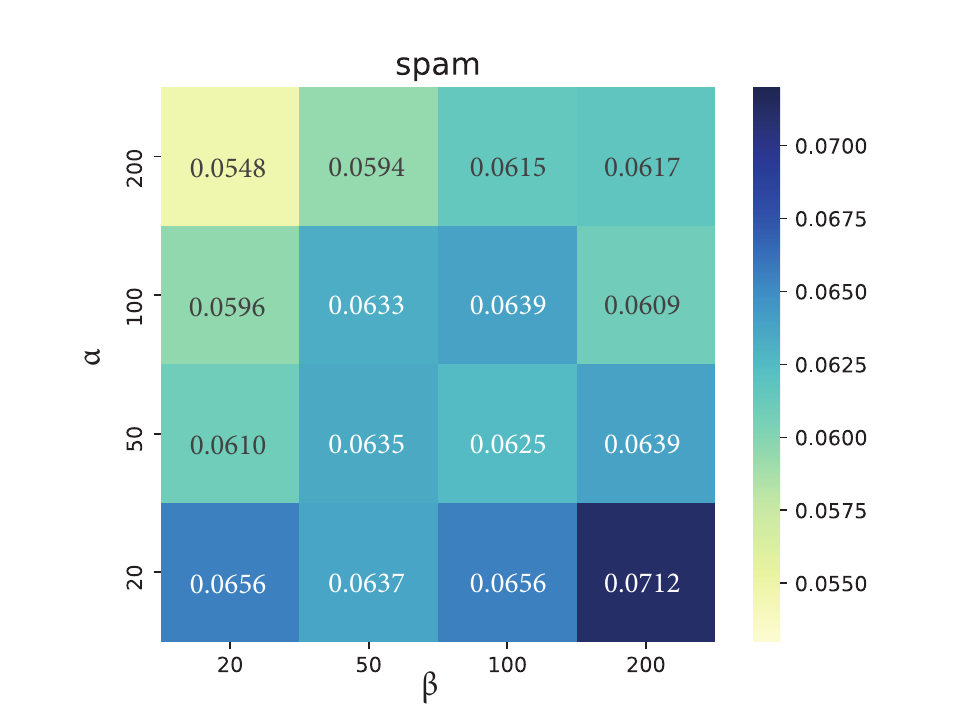}}}\hfill
\subfloat[WineQuality]{\label{fig:mdright}{\includegraphics[width=0.4\textwidth]{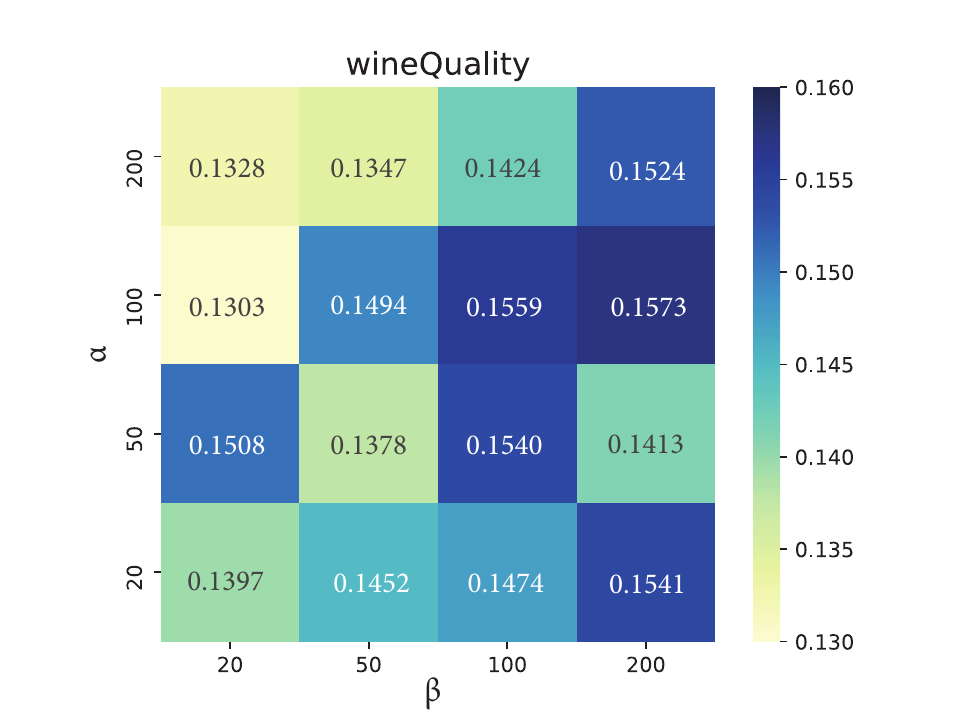}}}
\caption{ RMSE of PC-GAIN with various $\alpha$ and $\beta$.}
\label{WeightsEffects}
\end{figure}

First, we investigate the influences of the weight parameters $\alpha$ and $\beta$ in the objectives \eqref{PCGAINObj}.
We use a heat map manner to depict RMSE of PC-GAIN with various weights.
As shown in Figure \ref{WeightsEffects},
compared with $ \alpha $,
a relative smaller $ \beta $ can yield better results.
This implies that in  the framework of PC-GAIN,
the classifier loss has a greater impact on the imputation quality than the discriminator.

\begin{figure}[hbt]
\centering
\subfloat[News]{\label{fig:mdleft}{\includegraphics[width=0.333\textwidth]{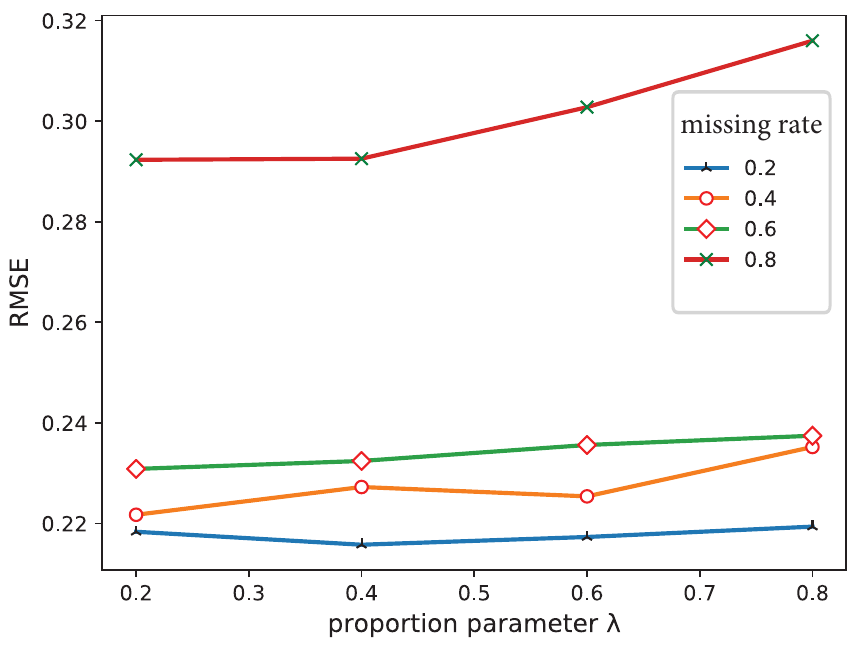}}}\hfill
\subfloat[Spam]{\label{fig:mdright}{\includegraphics[width=0.333\textwidth]{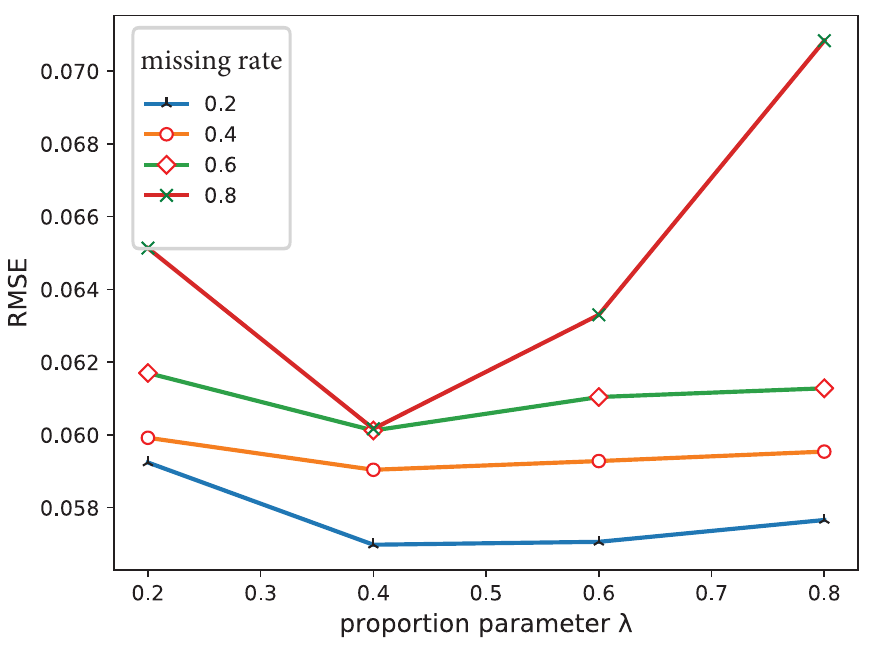}}}\hfill
\subfloat[WineQuality]{\label{fig:mdleft}{\includegraphics[width=0.333\textwidth]{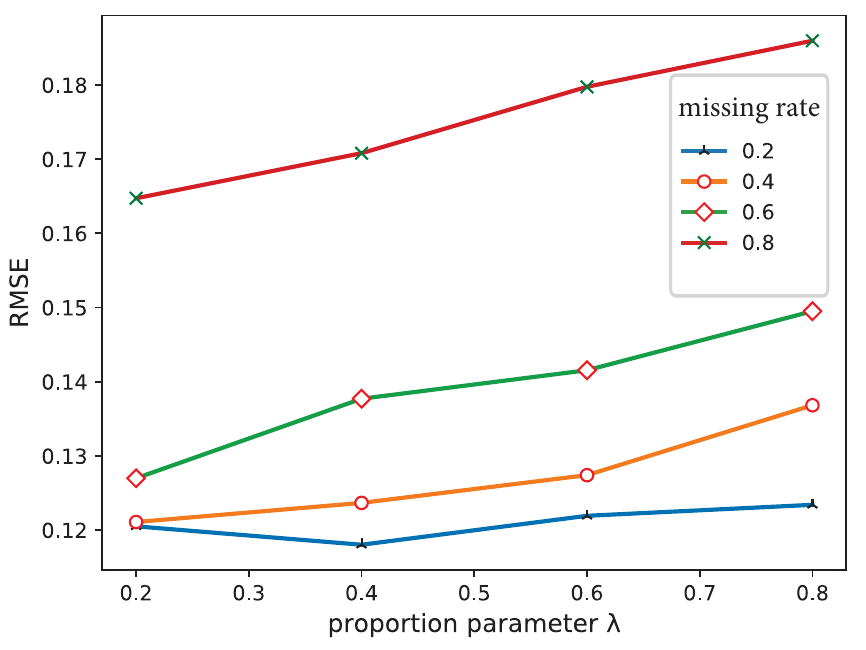}}}
\caption{ The influence of the proportion parameter $\lambda$.}
\label{lamda}
\end{figure}

\smallskip
In the pre-training stage, there is a proportion parameter $\lambda$, which controls the number of selected data in pre-training.
Next, we examine the effect of this parameter.

\smallskip
We can see from Figure \ref{lamda} that $ \lambda $  has a decisive effect on the quality of imputation.
The performance of PC-GAIN deteriorates rapidly when $\lambda$ becomes too large.
This is most likely because too many incomplete data may increase the unreliability of pseudo-labels,
 and thus reduces the quality of the imputation.
We also find that  the optimal choice of $\lambda$ for UCI datasets is usually below than 0.4.

\begin{figure}[H]
	\centering
	\scalebox{0.5}{\includegraphics{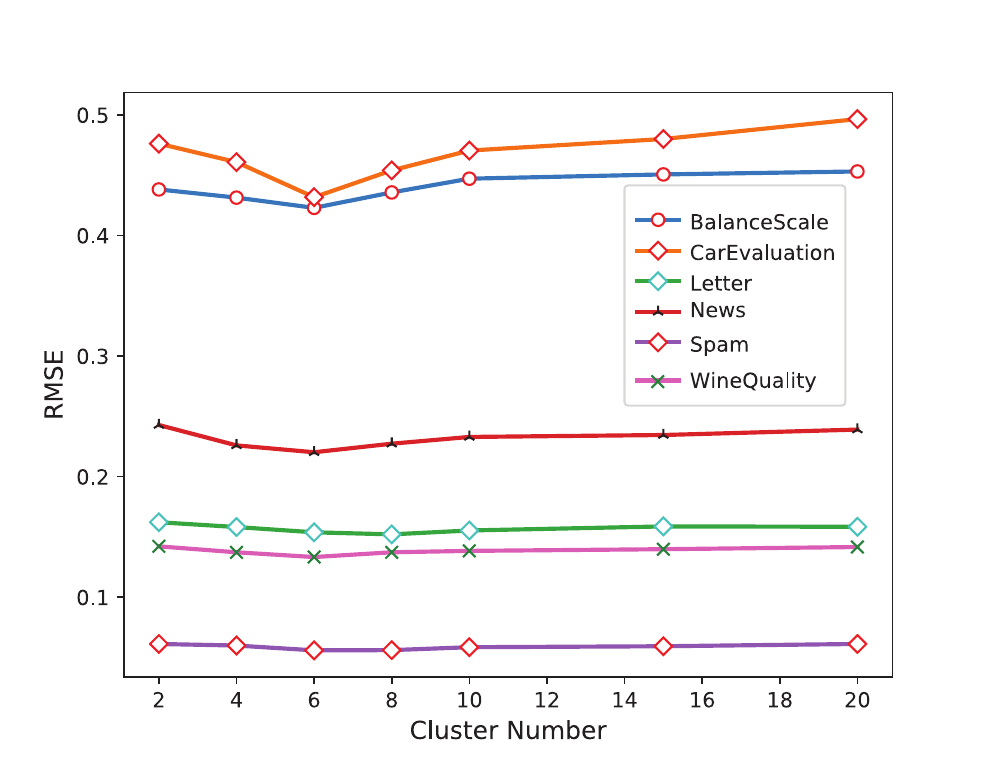}}
	\caption{RMSE of PC-GAIN  with various number of clusters.}
	\label{ablation_K}
\end{figure}

\smallskip
Finally, we examine the influence of the number of clusters.
From Figure \ref{ablation_K},
we can observe that the performance of PC-GAIN is stable with respect  to the number of clusters.
In most cases, PC-GAIN achieves a better results when $ K $ belonging 4 to 8, regardless of the true number of categories.
This property is very desirable,
because in practice we only need to consider a small  $ K $
even if the actual number of categories is large,
and thus save the computational overhead.

\subsection{Imputation Accuracy under Various Clustering and Classification Methods}
The proposed PC-GAIN utilized a clustering algorithm and an auxiliary classifier to improve the GAIN.
In this section, we analyze the relationship between the final imputation results and the classification and clustering methods used.

The classification methods adopted in the analysis are as follows:
Multiclass SVM \cite{Hsu2002} (has high classification accuracy),
a fully-trained neural network classifier (has high classification accuracy),
a little-trained neural network classifier (has low classification accuracy),
and a no-trained neural network classifier(completely random classification).
The clustering methods used in this analysis are:
KM(KMeans) \cite{Hartigan1979ak}, SC (SpectralClustering) \cite{Von2007tutorial}, KMPP(KMeans++) \cite{Arthur2007},
and AC(AgglomerativeClustering) \cite{Murtagh2012algorithms}.
Two quantitative indicators Ca (Calinski Harabasz) \cite{Mau2002} and Si (Silhouette Coefficient) \cite{Rous1987}
are used  to measure the quality of clustering.
The higher the value of these indicators, the better.

Table \ref{ClassifierAndCluster} shows the imputation results
under various clustering and classification methods
on two toy datasets and a realistic UCI dataset,
where the baseline denotes the result of the original GAIN.
 \textbf{Square} is a two-dimensional toy data with four categories, and the data of each category is strip  distribution \cite{Jain1999}.
\textbf{Twomoons} is a classic double moon shape data \cite{Zhou2004}.
\textbf{News} comes from the UCI dataset. RMSE is used to evaluate the imputation results. The lower the RMSE value, the better.

\begin{table}[!htb]
\center
\caption{{Comparison of imputation results under various clustering and classification methods.
}}
\label{ClassifierAndCluster}
\scalebox{0.68}
{\begin{tabular}{|c|c|c|c||c|c|}
	\hline
	{\textbf{Dataset}}	&\diagbox{{Cluster}}{{Classifier}}		&{Multiclass-SVM	} &{NN(complete train)}	&{NN(moderate train)	} &{NN(not train)}		\\
\hline
				&{Baseline}						&\multicolumn{4}{|c|}{{0.3545}}									\\
\cline{2-6}
				&{KM(Ca:228.4733 , Si:0.5778)}		&{0.2611}	&{0.2657}		&{0.3277}		&{0.3580}		\\
	{Square}		&{SC(Ca:241.5169 , Si:0.6295)}		&{0.2454}	&{0.2512	}	&{0.3143}		&{0.3538	}	\\
				&{KMPP(Ca:231.5886 , Si:0.5862)}		&{0.2528}	&{0.2530}		&{0.3017	}	&{0.3497}		\\
				&{AC(Ca:234.0248 , Si:0.6024)}		&{0.2577}	&{0.2615}		&{0.3225	}	&{0.3585}		\\
\hline
				&{Baseline}						&\multicolumn{4}{|c|}{{0.3230}}									\\
\cline{2-6}
				&{KM(Ca:437.2172 , Si:0.4808)}		&{0.2592}	&{0.2591	}	&{0.2892}		&{0.3227}		\\
	{Twomoons}		&{SC(Ca:422.9569 , Si:0.4697)}		&{0.2506}	&{0.2405	}	&{0.2749}		&{0.3308}		\\
				&{KMPP(Ca:361.1315 , Si:0.4355)}		&{0.2501}	&{0.2499}		&{0.2908}		&{0.3172}		\\
				&{AC(Ca:410.3813 , Si:0.4586)}		&{0.2583}	&{0.2510	}	&{0.2738	}	&{0.3099}		\\
\hline
				&{Baseline	}					&\multicolumn{4}{|c|}{{0.2532}}									\\
\cline{2-6}
				&{KM(Ca:259.7938 , Si:0.2165)}		&{0.2209}	&{0.2269}		&{0.2340}		&{0.2774}		\\
	{News}			&{SC(Ca:244.6607 , Si:0.2136)}		&{0.2228}	&{0.2253}		&{0.2297	}	&{0.2674	}	\\
				&{KMPP(Ca:194.9008 , Si:0.1829)}		&{0.2256}	&{0.2269}		&{0.2390	}	&{0.2705	}	\\
				&{AC(Ca:190.7064 , Si:0.1936)}		&{0.2222}	&{0.2227}		&{0.2358	}	&{0.2644}		\\
\hline
\end{tabular}}
\end{table}

Each column in the Table \ref{ClassifierAndCluster} reflects the imputation results of different clustering methods and a certain classification method, and each row in the Table \ref{ClassifierAndCluster} reflects the imputation results of a certain clustering method and different classification methods. As can be seen from Table \ref{ClassifierAndCluster} (all rows and the first two columns), the model has strong robustness for various clustering methods and classification methods on each dataset.
Although different clustering methods have different effects on a dataset,
when the clustering results are fully learned by the classifiers, they can be used to guide the network to impute the missing data. As can be seen from the last three columns of the Table \ref{ClassifierAndCluster},
when the classifier has not been fully trained, the classification accuracy is low, and the classifier's supervisory role on the imputation network is weak at this time.
However, with the continuous training of the classifier, the classification accuracy increases,
and the ability of the classifier to supervise the imputation network is also gradually enhanced,
so that the imputation network can output more reliable imputed results.

\subsection{Image Inpainting}
In this section,
we show that PC-GAIN  can also improve the performance of GAIN in image inpainting task.
To this end, we consider the MNIST dataset \cite{LeCun2010}.
For each image in MNIST, we remove 50\% and 80\% pixels uniformly at random.
We compare the imputation results using FID index \cite{Heusel2017}.
FID represents the distance between the eigenvectors of the generated image and the eigenvectors of the real image,
 a lower FID value often means that the generated image is closer to the real image.

\begin{figure}[H]
\centering
\subfloat[Incomplete image]{\label{fig:mdleft}{\includegraphics[width=0.33\textwidth]{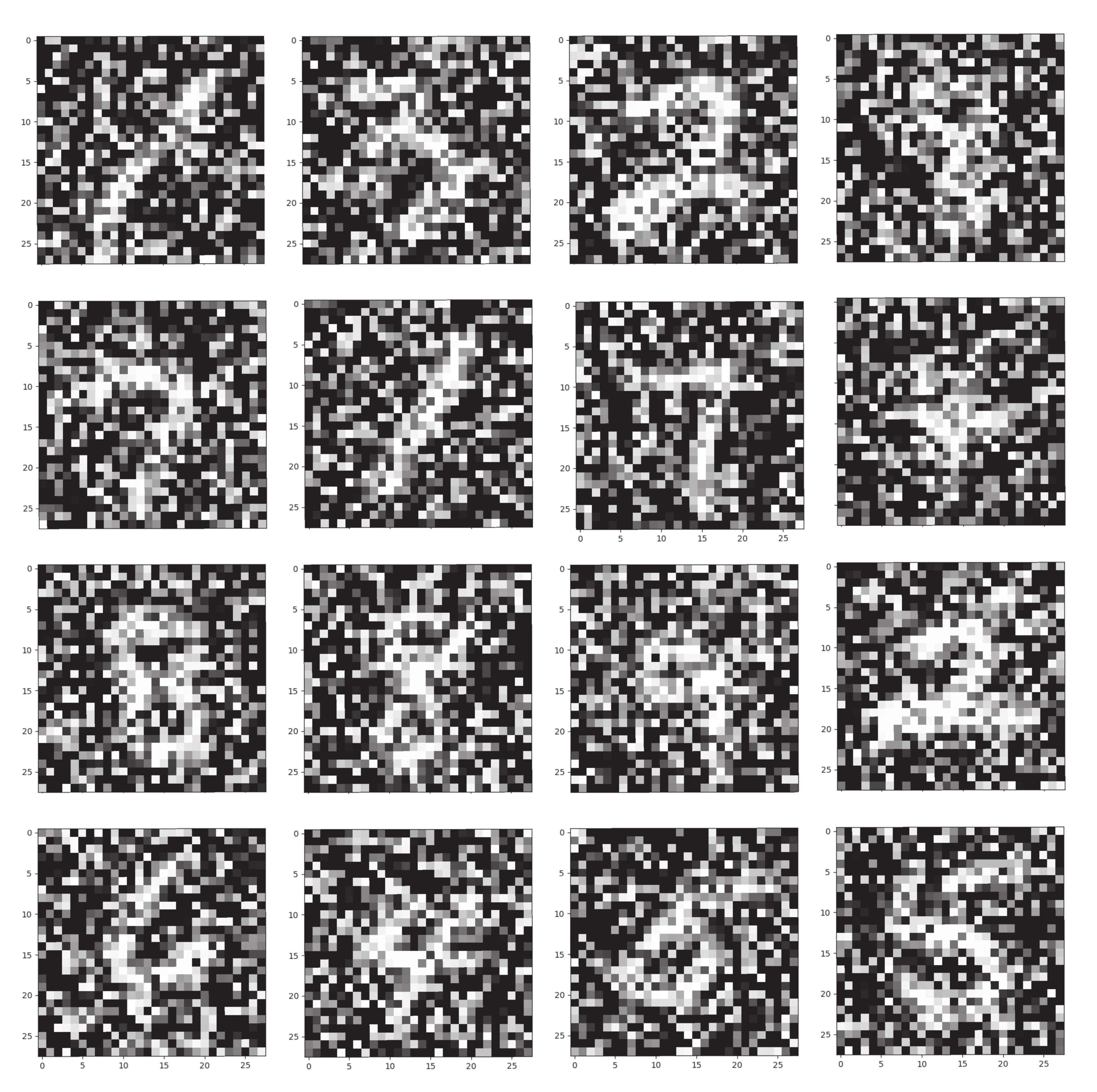}}}\hfill
\subfloat[GAIN (FID 1.721)]{\label{fig:mdright}{\includegraphics[width=0.33\textwidth]{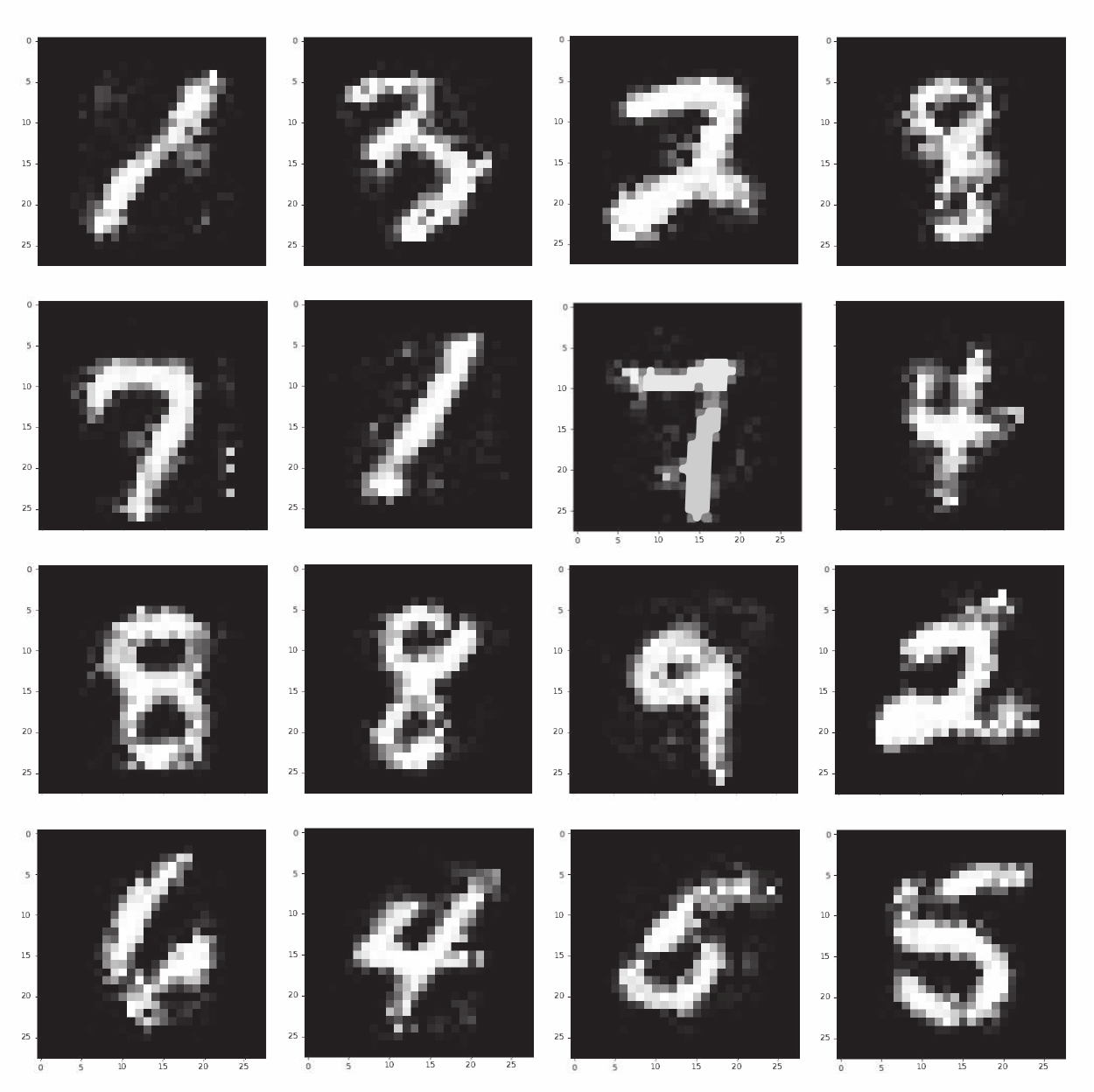}}}\hfill
\subfloat[PC-GAIN (FID 1.551)]{\label{fig:mdleft}{\includegraphics[width=0.33\textwidth]{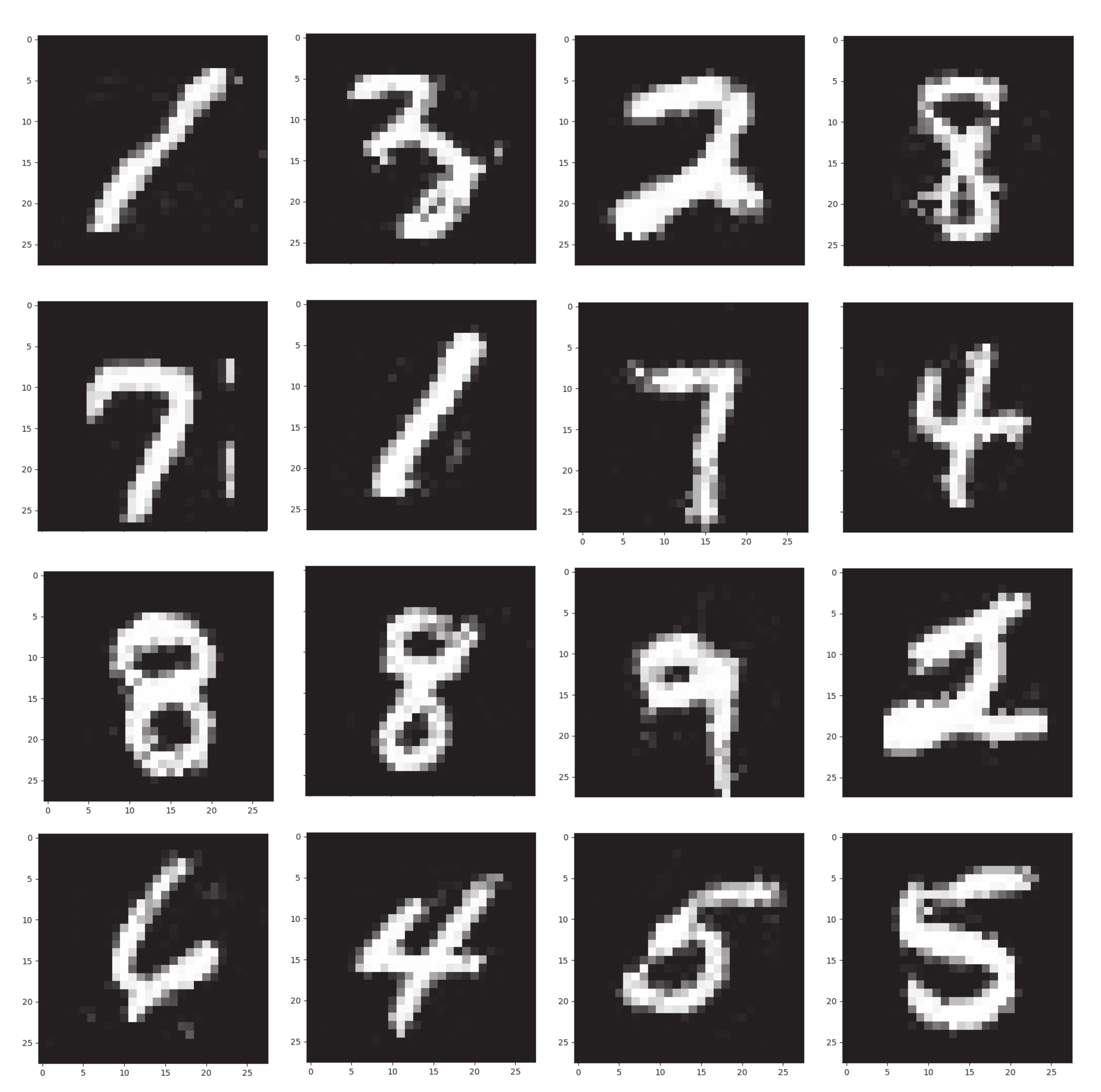}}}
\caption{ Imputed images on MNIST under a 50\% missing rate}
\label{MNIST50}
\end{figure}

\begin{figure}[H]
\centering
\subfloat[Incomplete image]{\label{fig:mdleft}{\includegraphics[width=0.33\textwidth]{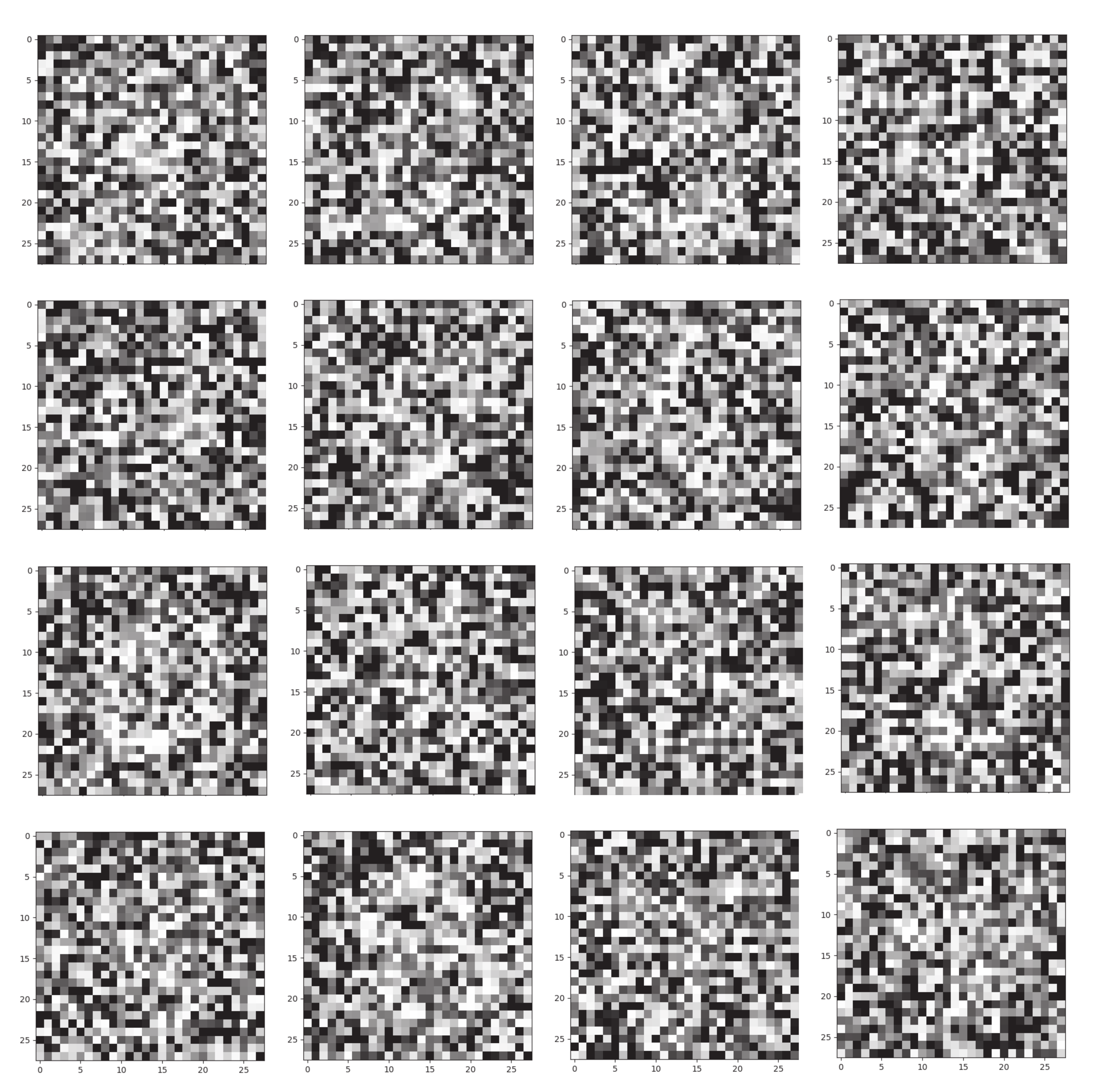}}}\hfill
\subfloat[GAIN (FID 5.642)]{\label{fig:mdright}{\includegraphics[width=0.33\textwidth]{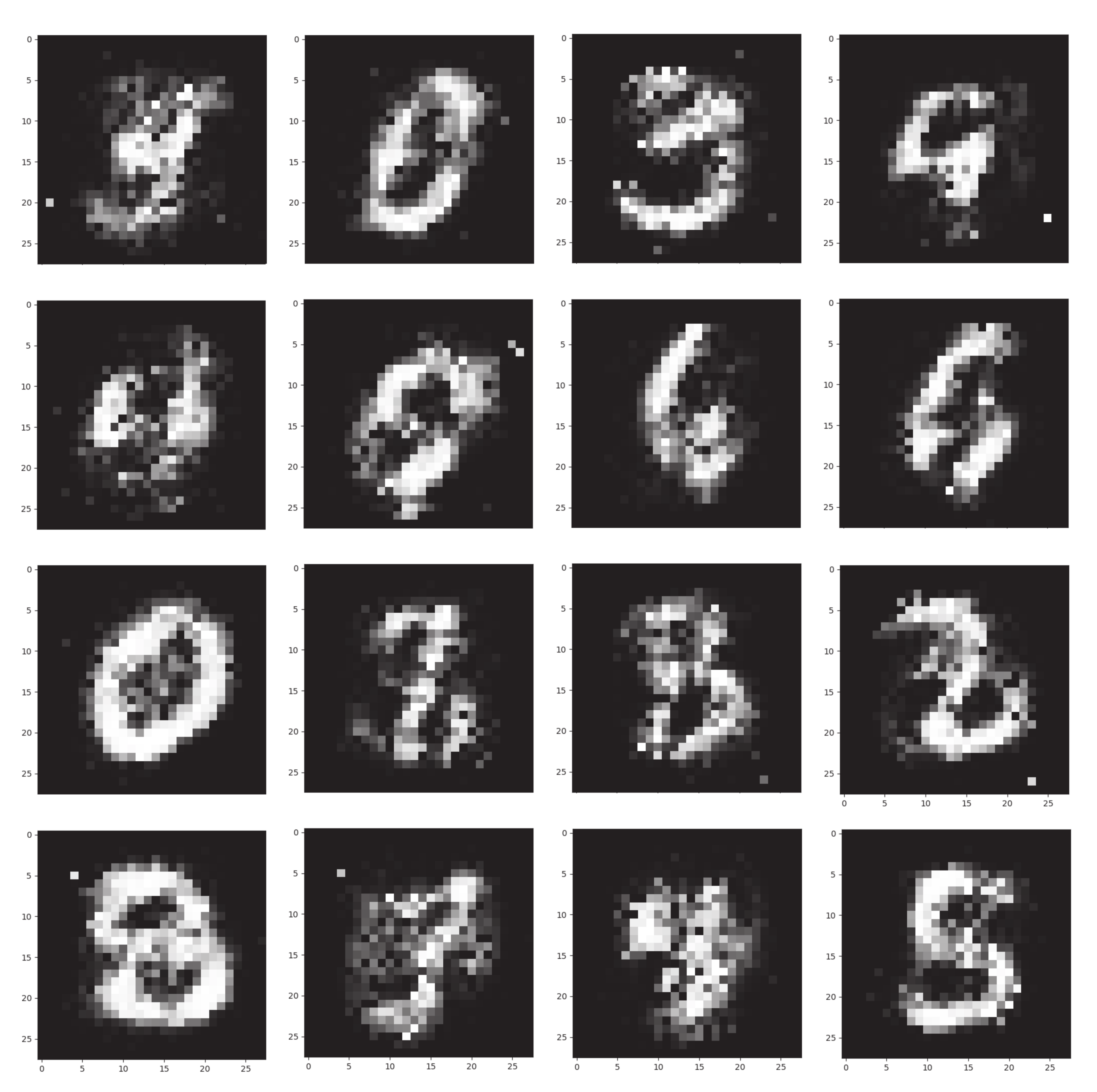}}}\hfill
\subfloat[PC-GAIN(FID 4.520)]{\label{fig:mdleft}{\includegraphics[width=0.33\textwidth]{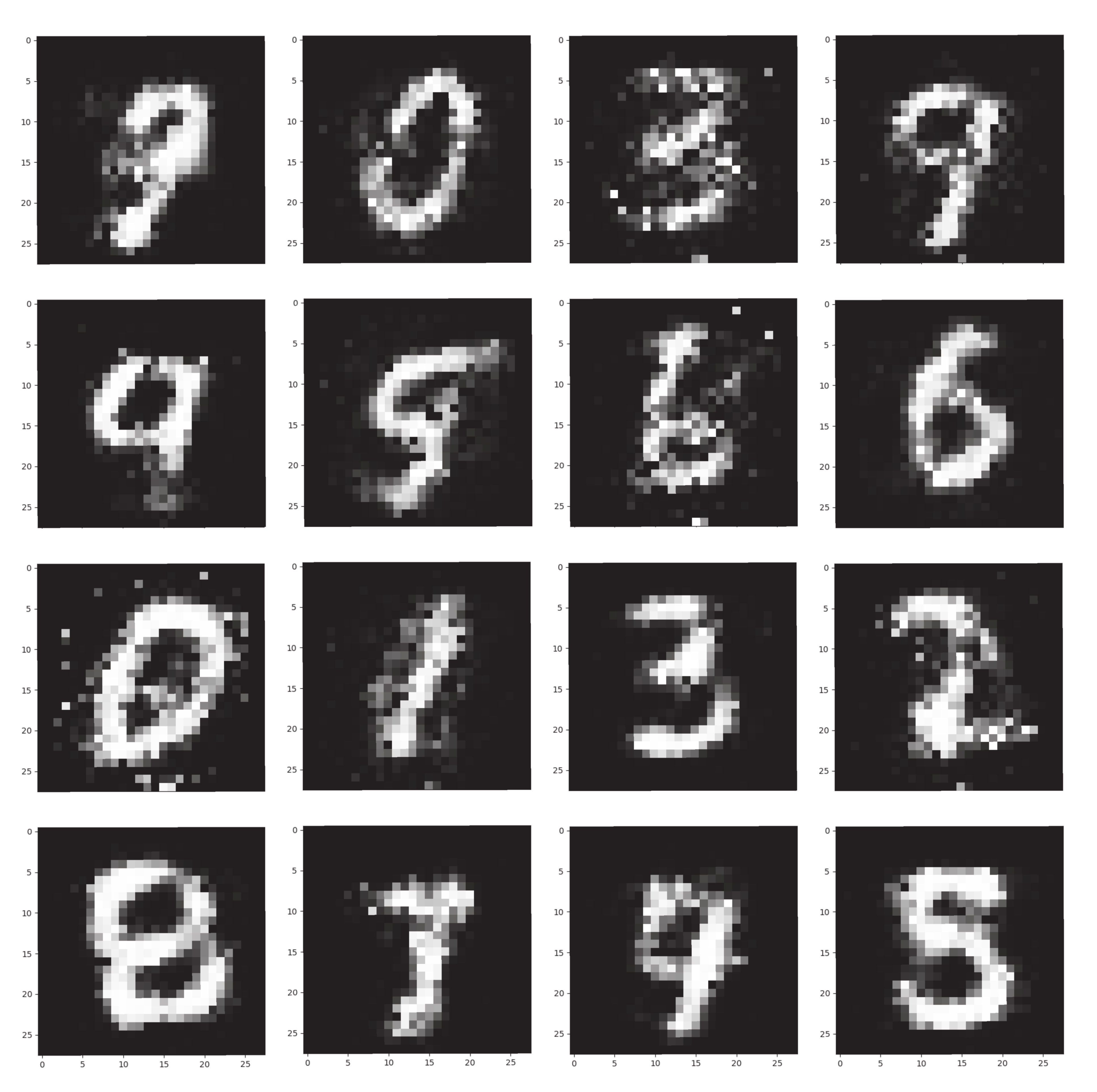}}}
\caption{Imputed images on MNIST under  a 80\% missing rate}
\label{MNIST80}
\end{figure}

It is obvious from Figures \ref{MNIST50} and \ref{MNIST80} that
the imputed images of  PC-GAIN  are more cohesive and smoother than those of GAIN.
Moreover, the output of PC-GAIN has a lower FID value at both cases,
and the advantage becomes greater at a higher miss rate.

\section{Concluding Remarks}
Based on the recent work of \cite{Yoon2018},
we propose a novel generative model called PC-GAIN for missing data imputation.
With the aid of an auxiliary classifier, which has been pre-trained using a subset of low-missing-rate samples and the corresponding pseudo-labels,
the generator tries to produce indistinguishable  imputation results that have obvious categorical characteristics.
It is worth noting that this classifier is always fixed during the training of generative adversarial networks, so the proposed method is easy to implement.
On the whole, PC-GAIN can be regarded as an improved version of GAIN \cite{Yoon2018},
which promotes the performance of the original model significantly without using any supervision, especially under a high missing rate.

\smallskip
The key of PC-GAIN  is to exploit the potential category information contained in the missing data to enhance the imputation results.
This novel idea is very general and can be applied to other existing frameworks,
as long as the potential category information can be accurately captured.
However, compared with traditional and other deep learning methods,
an additional pre-training step is needed in our framework,
which requires more time in training process and has a larger number of hyperparameters to be adjusted.

\smallskip
Although we only focus on the MCAR case in this paper,
our approach can be extended to handle both MAR and MNAR mechanisms.
Note that the selected pre-training subset has an important impact on the quality of the pseudo-labels, and thereby affects the imputation results.
For MAR, we can still pre-train some low-missing-rate data to ensure the quality of the pseudo-labels.
However, for MNAR,  it is not a good idea to simply select a subset with low missingness,
which may lead to selection bias and reduce the accuracy of the pseudo-labels.
A more suitable pre-training subset should be chosen according to the corresponding missing mechanism.
It will be a meaningful work to apply the main idea of PC-GAIN in other frameworks to deal with MNAR case.

\bigskip
\textbf{Acknowledgments}. This research is partially supported by National Natural Science Foundation of China (11771257) and Natural Science Foundation of Shandong Province (ZR2018MA008)

\end{document}